\pdfoutput=1
\documentclass[11pt]{article}

\usepackage{ACL2023}

\usepackage{times}
\usepackage{latexsym}

\usepackage[T1]{fontenc}
\usepackage[utf8]{inputenc}

\usepackage{microtype}
\usepackage{amsfonts}
\usepackage{colortbl}
\usepackage{inconsolata}
\usepackage{graphicx}
\usepackage{amsmath}
\usepackage{booktabs}
\usepackage{multirow}
\usepackage{multicol}
\newcommand*{\SmallImage}[1]{\includegraphics[width=0.32cm,height=!,valign=m]{#1}}

\newcommand*{\Image}[1]{\includegraphics[width=0.4cm,height=!,valign=m]{#1}}%
\usepackage{inconsolata}
\usepackage{capt-of}
\usepackage{tabularx}
\usepackage{booktabs}
\usepackage{scalerel}
\usepackage{listings}
\usepackage[export]{adjustbox}
\usepackage{pifont} 
\definecolor{mygreen}{RGB}{59,205,62}

\newcommand{\green}[1]{\textcolor{mygreen}{{#1}}}

\usepackage{adjustbox}

\makeatother

\title{PaCE: Unified Multi-modal Dialogue Pre-training with \\
Progressive and Compositional Experts}

\author{Yunshui Li$^{1,2}$\footnotemark[1]\ \footnotemark[2] \quad  Binyuan Hui$^{3}$\footnotemark[1] \quad Zhichao Yin$^{1,4}$ \ \   Min Yang$^{1}$\footnotemark[3] \quad \textbf{Fei Huang}$^{3}$ \ \  \textbf{Yongbin Li}$^{3}$\footnotemark[3] \\
        $^{1}$Shenzhen Institute of Advanced Technology, Chinese Academy of Sciences \\
        $^{2}$University of Chinese Academy of Sciences \\
        $^{3}$DAMO Academy, Alibaba Group\\
        $^{4}$University of Science and Technology of China\\
        \texttt{\{ys.li, min.yang\}@siat.ac.cn, \{binyuan.hby, shuide.lyb\}@alibaba-inc.com}\\
        \small\url{https://github.com/AlibabaResearch/DAMO-ConvAI/tree/main/pace}
        }

\begin{document}
\maketitle

\renewcommand{\thefootnote}{\fnsymbol{footnote}}
\footnotetext[1]{Equal contribution.}
\footnotetext[2]{Work done during an intern at Alibaba DAMO Academy.}
\footnotetext[3]{Corresponding author.}

\renewcommand{\thefootnote}{\arabic{footnote}}

\begin{abstract}
Perceiving multi-modal information and fulfilling dialogues with humans is a long-term goal of artificial intelligence. Pre-training is commonly regarded as an effective approach for multi-modal dialogue. However, due to the limited availability of multi-modal dialogue data, there is still scarce research on multi-modal dialogue pre-training. Yet another intriguing challenge emerges from the encompassing nature of multi-modal dialogue, which involves various modalities and tasks. Moreover, new forms of tasks may arise at unpredictable points in the future. Hence, it is essential for designed multi-modal dialogue models to possess sufficient flexibility to adapt to such scenarios. 
This paper proposes \textbf{PaCE}, a unified, structured, compositional multi-modal dialogue pre-training framework. It utilizes a combination of several fundamental experts to accommodate multiple dialogue-related tasks and can be pre-trained using limited dialogue and extensive non-dialogue multi-modal data. Furthermore, we propose a progressive training method where old experts from the past can assist new experts, facilitating the expansion of their capabilities. 
Experimental results demonstrate that PaCE achieves state-of-the-art results on eight multi-modal dialog benchmarks.

\end{abstract}

\section{Introduction}

Enabling seamless communication between humans and machines is a long-standing goal of artificial intelligence research.
The recent emergence of chatGPT~\footnote{https://chat.openai.com/} has increased confidence in the potential for achieving this goal. 
Beyond the use of textual language as a unique interface between humans and machines, perceiving and utilizing multi-modal information, especially visual information, has become a crucial capability known as multi-modal dialogue~\citep{shuster2020multi,sun2021multimodal}.

\begin{figure}[t]
    \small
    \centering
    \includegraphics[width=0.95\columnwidth]{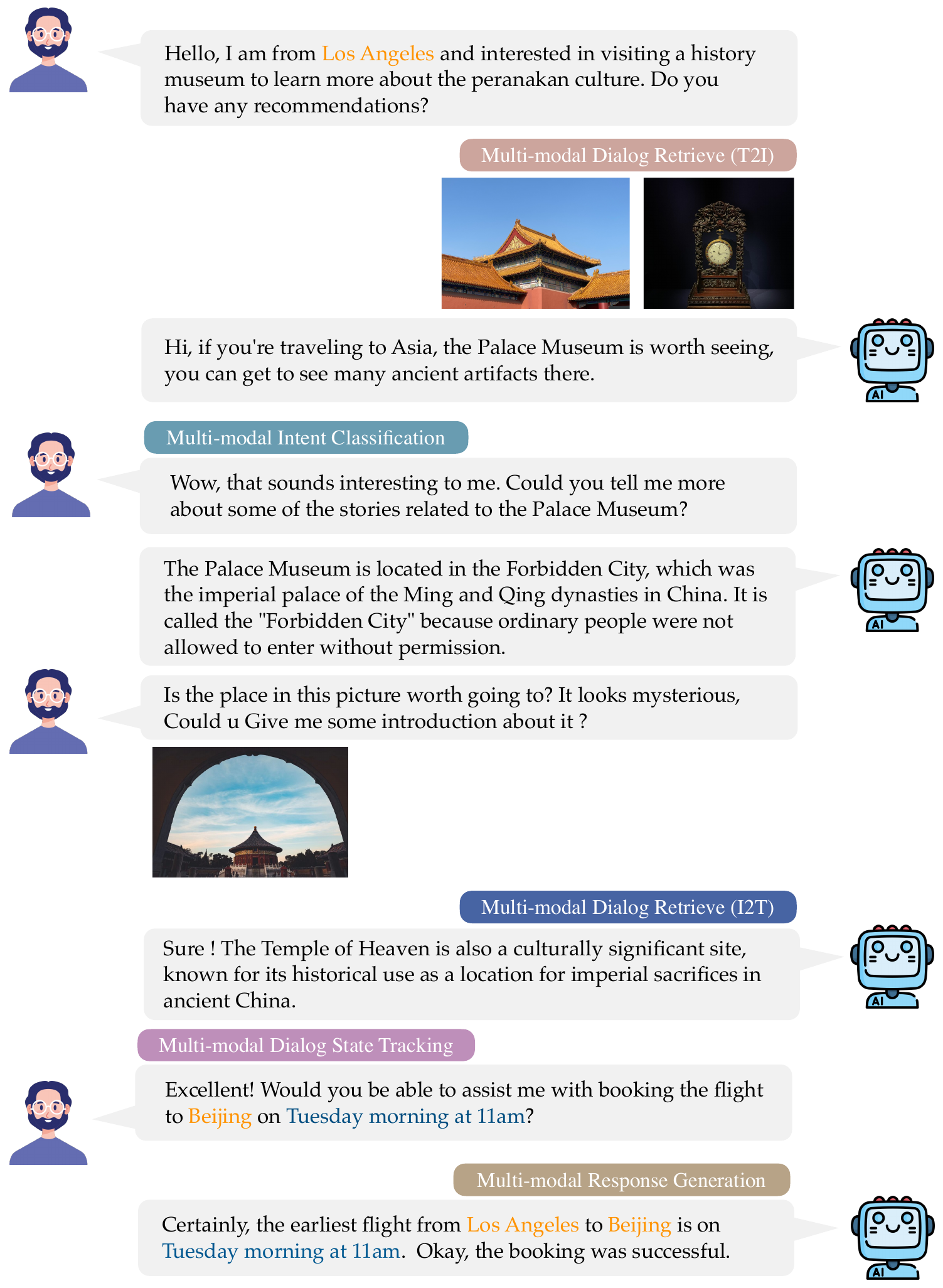}
    \caption{An example of multi-modal dialogue, which involves multiple tasks, including multi-modal intent classification, multi-modal state tracking, multi-modal dialog retrieval and response generation.}
    \label{intro}
\end{figure}

To facilitate the research on multi-modal dialogue, plenty of specific tasks and datasets have emerged in the community~\citep{das2017visual,shuster2018image,feng2022mmdialog,long2023spring}. However, the overall quantity of data is still limited. 
Furthermore, multi-modal dialogue presents a greater challenge compared to traditional text-only dialogue track~\citep{hui2021dynamic,he2022galaxy,si2022mining}, as it involves the integration of various modalities and more intricate task scenarios. As shown in Figure~\ref{intro}, the central tasks of multi-modal dialogue include multi-modal intent classification~\citep{zang2021photochat}, multi-modal dialogue retrieval~\citep{das2017visual,zang2021photochat}, multi-modal dialogue state tracking~\citep{liao2021mmconv}, and multi-modal response generation~\citep{kottur2021simmc}. 
Despite pre-training having become the consensus for multi-task learning in machine learning~\citep{devlin2018bert, radford2019language,radford2021learning}, the research on pre-training models for multi-modal dialogue is an area that is yet to be fully explored.

In this paper, we focus on building pre-trained models of multi-modal dialogue. 
A key challenge is to unify different modalities and task forms, and make the best use of existing multi-modal dialog and non-dialog data. A recent popular trend on textual tasks is to build unified pre-trained foundation models by multi-task learning, e.g., T5~\citep{raffel2020exploring}. However, it attempts to mix all tasks learned from scratch thus is difficult to control the learning process, which is a completely black box. Although the Mixture-of-Experts (MoE)~\citep{fedus2021switch,du2022glam} architecture attempts to select independent experts for each input sample through token-level routing, it lacks specific semantics, i.e., it is entirely unknown what the experts are responsible for.
We hope to find a new way to handle many multi-modal dialog tasks simultaneously and combine existing concrete skills to learn new tasks more efficiently.

To this end, we propose \textbf{PaCE}, a unified multi-modal dialogue pre-training framework with
\textbf{P}rogressive \textbf{a}nd \textbf{C}ompositional \textbf{E}xperts. 
\textbf{First}, we decompose complicated multi-modal dialogue into fundamental sub-capabilities that could be learned with specific data.
Different from traditional MoE, each expert in PaCE is tailored to one specific fundamental sub-capability of multi-modal dialogue, including \textsc{Caption}, \textsc{Context}, \textsc{Image}, \textsc{Grounding} and \textsc{Generation}.
% Different from traditional MoE, each expert in PaCE is tailored to one specific fundamental sub-capability of multi-modal dialogue.  
\textbf{Second}, we propose a progressive pre-training strategy to evolve the model by controlling the combination of experts in different pre-training phases. 
Specifically, in stage I, we first train on multi-modal non-dialogue data to obtain \textsc{Caption}, \textsc{Image}, and \textsc{Grounding} experts. In stage II, we train the \textsc{Context} expert, which is guided by the \textsc{Caption} expert on multi-modal dialog data to learn the dependencies in context. 
Furthermore, a dialogue \textsc{Generation} expert is derived by adding a response generation task based on the previously learned experts. 
\textbf{Third}, for pre-training PaCE, we collect a multi-modal dialog corpus with 1.4 million dialogs and a multi-modal non-dialog corpus with 4 million samples. Once the pre-training of PaCE is finished, we can flexibly select different capability experts to solve a specific downstream task.

\begin{figure}[t!]
    \small
    \centering
    \includegraphics[width=\columnwidth]{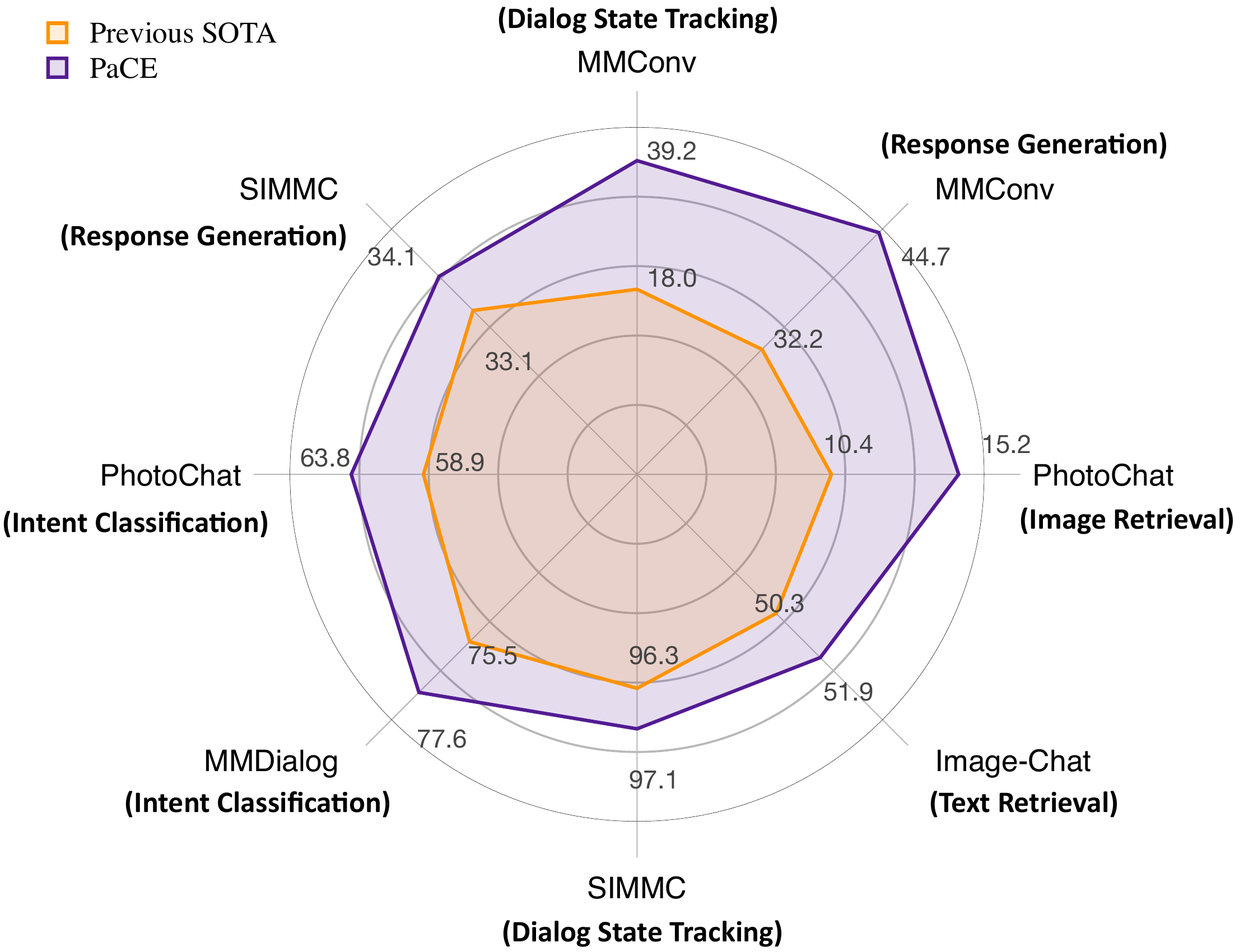}
    \caption{PaCE achieves state-of-the-art performances on a broad range of dialogue tasks compared with other customized or foundation models. }
    \label{fig:result}
\end{figure}

As illustrated in Figure~\ref{fig:result}, PaCE achieves state-of-the-art performance across a broad range of multi-modal dialogue benchmarks spanning four diverse downstream tasks, i.e., multi-modal intent classification, multi-modal dialogue retrieval, multi-modal state tracking, and multi-modal response generation This demonstrates that PaCE not only possesses a flexible model architecture but also exhibits adaptable training methodologies, resulting in remarkable performance.

% \clearpage
\section{Related Work}

\paragraph{Pre-trained Vision-Language Models}
% \liyunshui{
The pre-training paradigm, with its successes in natural language processing~\citep{devlin2018bert,radford2019language}, has sparked a revolution in Multi-modal Learning. ViLBERT~\citep{lu2019vilbert} was the first work to adapt the BERT-like architecture for visual-language modeling, allowing for learning joint representation of images and texts. ViLT~\citep{kim2021vilt} constructed the vision module in the same way as the text module with a unified Transformer~\citep{vaswani2017attention}, eliminating the need for resource-intensive image feature extraction and significantly accelerating the model. CLIP~\citep{radford2021learning} employed contrast learning to directly align images with natural language texts, eliminating the constraints of predefined image categories. ALIGN~\citep{jia2021scaling} and Florence~\citep{yuan2021florence} further generalized this idea on noisier but larger image-text pairs. These models have demonstrated the ability to learn strong image and text representations for cross-modal alignment tasks. In addition, a number of models~\citep{cho2021unifying,wang2021simvlm,wang2022unifying,yu2022coca,alayrac2022flamingo} employed auto-regressive models to model the association between images and texts, using a unified generation approach to construct the task in an end-to-end manner. Although pre-trained vision-language models have shown promising results, they mainly focus on caption texts which are intrinsically different from human conversations~\citep{kulhanek2021augpt}. To our best knowledge, the proposed PaCE model is the first multi-modal dialogue pre-training model. 

\paragraph{Multi-Modal Dialogue Modeling}
Numerous advanced works have been proposed along with the development of multi-modal dialogue datasets \cite{das2017visual,mostafazadeh2017image,shuster2018image,zang2021photochat,zheng2021mmchat,kottur2021simmc,liao2021mmconv,feng2022mmdialog}. Several dialogue modeling works~\citep{qi2020two,lee2021constructing} have been conducted to improve the performance of conversational agents in image-grounded dialogue. ~\citet{zang2021photochat} proposed a dual-encoder model that utilized object labels to encode image features so as to perform a dialogue-based image retrieval task. 
Afterward, researchers~\citep{yang2021open,chen2021learning} explored enriching textual expressions of generated dialogue responses through associative vision scenes. For textual response tasks, ~\citet{zheng2021mmchat} proposed a multi-modal dialogue generation model based on Seq2Seq architecture, which was proved to be superior to the textual Seq2Seq model. ~\citet{lee2022learning} proposed a joint multi-modal encoder-decoder model to incorporate visual inputs.
However, the above models have demonstrated success in specific sub-tasks with a particular dataset, which cannot meet the requirements of a wide range of multi-modal dialogue tasks. To address this challenge, we propose a unified multi-modal dialogue pre-training model based on a divide-and-conquer strategy, which can combine different experts to complete a series of tasks.

\begin{figure*}[t!]
    \small
    \centering
    \includegraphics[width=\textwidth]{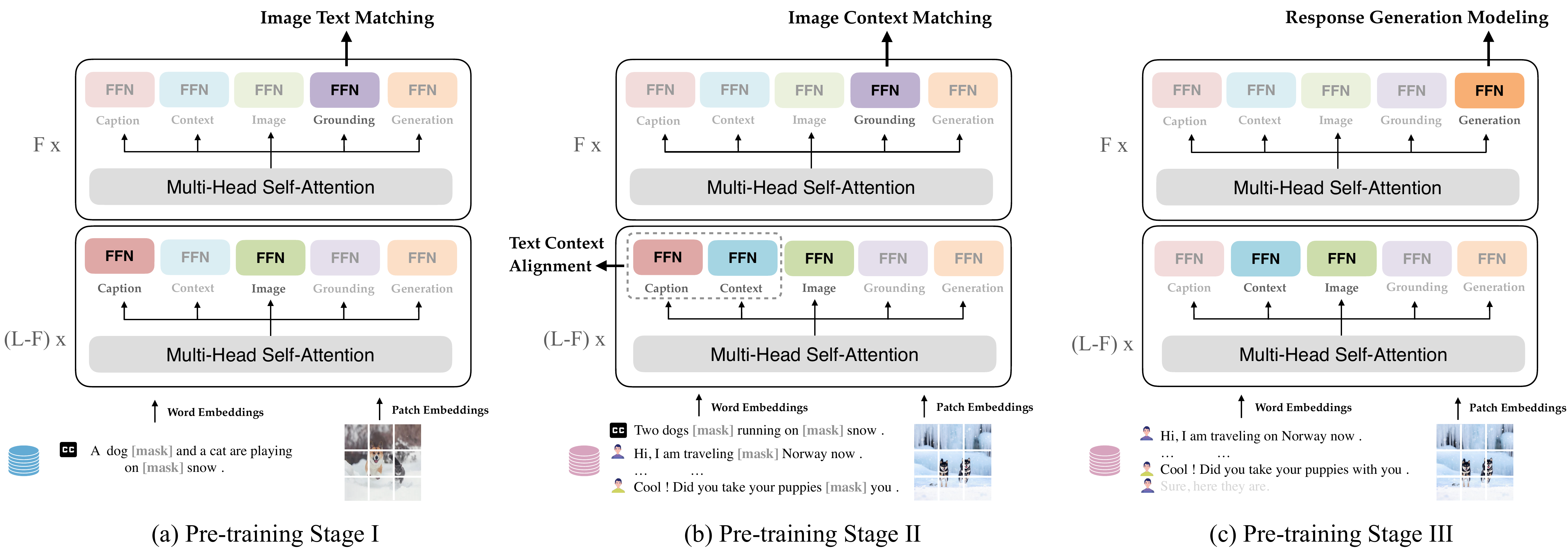}
    \caption{Three-stage training based on different combinations of experts, where the \Image{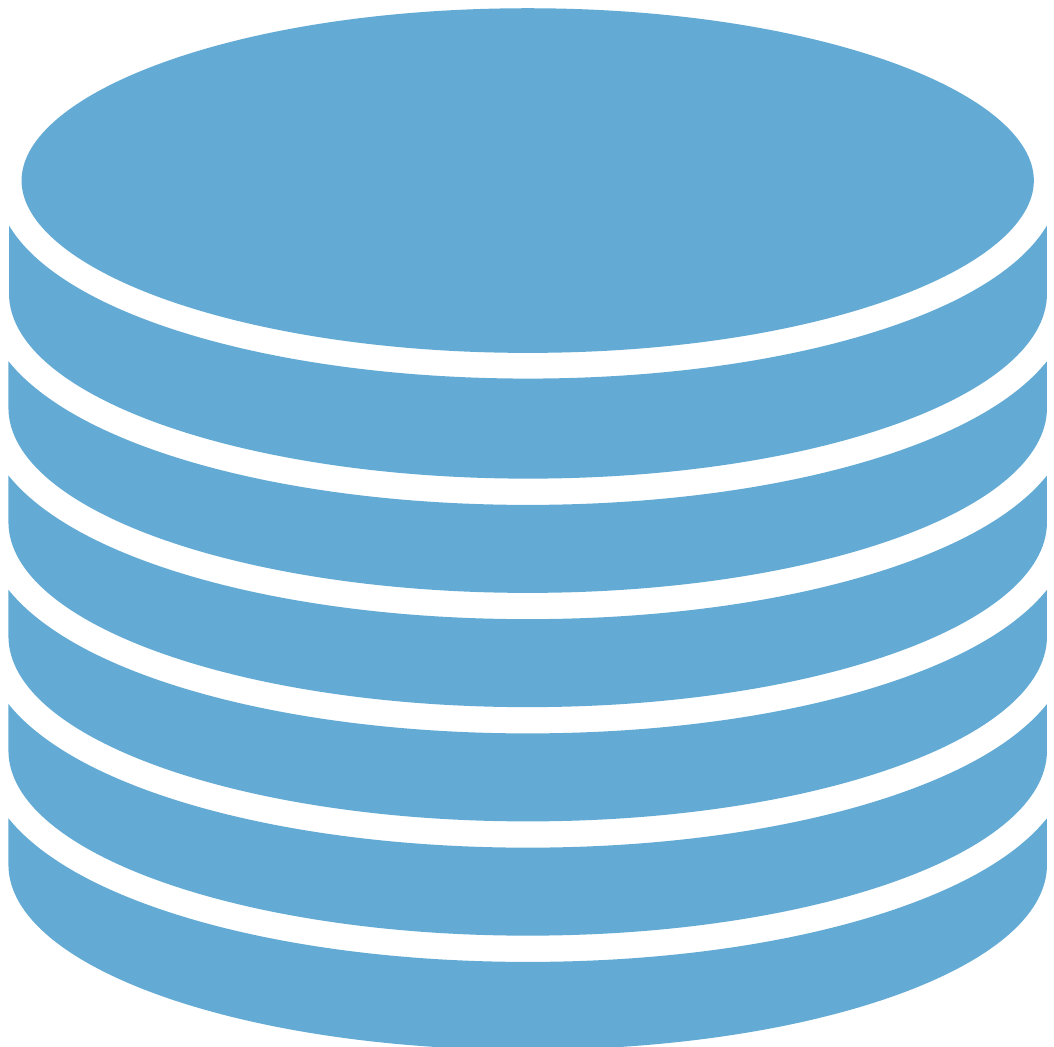} represents  multi-modal non-dialog data and works mainly in the first stage, while the \Image{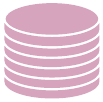} represents multi-modal dialog  data and works in the second and third stages. The \SmallImage{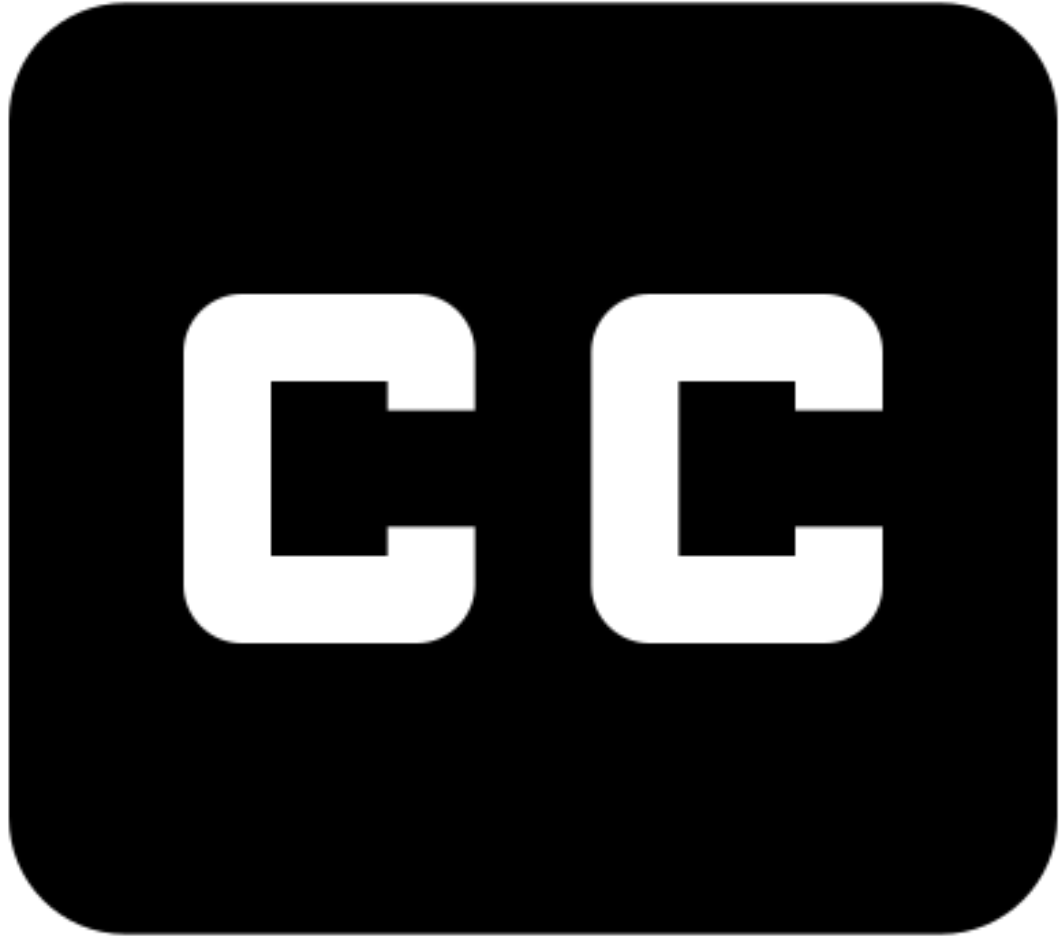} represents the caption of the input image.} 
    \label{method}
\end{figure*}

\section{Pre-training Data Construction}
In this paper, we collect both multi-modal non-dialogue and 
multi-modal dialogue data for PaCE pre-training.
The total statistics of our collected pre-training corpora are shown in Table \ref{data}. 
% TODO：云水补充相关细节
\paragraph{Multi-modal Non-dialogue Data (MultiNonDialog)}
Similar to previous work~\citep{kim2021vilt}, we first collect four multi-model non-dialogue datasets for image and text representation learning, including MSCOCO~\citep{lin2014microsoft}, VG~\citep{krishna2017visual}, SBU~\cite{ordonez2011im2text} and GCC~\citep{sharma2018conceptual}. In MultiNonDialog, each image is accompanied by one or more captions whose lengths are generally constrained to 20 tokens. Since GCC and SBU provide only image URLs, we collect the images via the given URLs which are still accessible. 

\begin{table}[t!]
    \centering
    \resizebox{0.48\textwidth}{!}{
        \begin{tabular}{clccl}
            % \toprule
            \toprule
            \textbf{Category} & \textbf{Dataset} & \textbf{Turns} & \textbf{Dialogs} & \textbf{Images} \\
            \cmidrule(lr){1-5}
            % \hline
            \multirow{4}{*}{\Image{icon/icon3.pdf} 
            MultiNonDialog} & CC3M  & 3.01M  & - & 3.01M\\
            & SBU & 867K  & -  & 867K \\
            & MSCOCO & 113K  & - & 567K \\
            & VG & 108K& - & 5.41M  \\
            \cmidrule(lr){1-5}
            \multirow{6}{*}{\Image{icon/icon2.pdf} MultiDialog}
            & VisDial & 1.2M & 120K  & 120K  \\
            & Image-Chat  & 400K & 202K & 202K \\
            & PhotoChat & 97.6K & 12.2K & 11K \\
            & MMConv  & 39.7K  & 5.1K  & 114K \\
            & SIMMC2.0 & 117K & 11K  & 1.5K \\
            & MMDialog & 4.82M & 1.08M & 1.53M \\

            \bottomrule
        \end{tabular}}
    \caption{\label{pretrain_data}
        Statistics of our collected pre-training corpora.
    }
    \label{data}
\end{table}
\paragraph{Multi-modal Dialogue Data (MultiDialog)}
We collect six existing multi-modal conversation corpora ranging from online forum chatting logs~\citep{das2017visual,shuster2018image, zang2021photochat, feng2022mmdialog} to customer service conversations~\citep{liao2021mmconv,kottur2021simmc} and build a large-scale multi-modal dialogue corpus. To ensure that each conversation has at least one corresponding image, we eliminate the text-only conversations from the original datasets. In addition, to satisfy the requirements of the Stage II pre-training, we use 
 the BLIP model~\citep{li2022blip} implemented by ~\citet{li2022lavis} to generate the appropriate textual caption for each image. The captions are  constrained to 20 tokens.

% 与image-caption数据分布上相对统一且数据量巨大不同，多模态对话数据集之间常常存在较大的差异且数量远远不足
\section{Pre-training Method}
Given a set of $n$ multi-modal dialogue samples $\mathcal{D}=\left\{\left(U_i, R_i\right)\right\}_{i=1}^n$, where $U_i$ and $R_i$ represent the dialogue context and response, respectively.
Compared to traditional textual dialogue, both $U_i=\left\{u_k^m\right\}_{k=1}^K$ and $R_i=\left\{r_q^m\right\}_{q=1}^Q$ can incorporate various types of information including textual texts and visual images, where $K$ and $Q$ are the number of elements, and $m \in \{t, v\}$ denotes the modality of  $U_i$ (or $R_i$). The notation $t$ indicates textual utterances, while $v$ indicates visual images.

% We choose Transformer as our backbone model for dialog modeling. 
We devise a divide-and-conquer pre-training strategy for multi-modal dialogue. Concretely, we decompose complicated multi-modal dialogue into five fundamental sub-capabilities and design five corresponding experts (i.e., \textsc{Caption}, \textsc{Context}, \textsc{Image}, \textsc{Grounding}, and \textsc{Generation} experts). Then, we propose a progressive training strategy to evolve the model by controlling the combination of experts in different pre-training phases. 
Next, we describe the input representation learning module, the divide-and-conquer pre-training strategy, the pre-training objectives, and the fine-tuning process in detail.

\subsection{Input Representation Learning}
The proposed model is designed to handle input data from two modalities: visual representations and textual representations. 
%We describe the details of each modality separately here.

\paragraph{Visual Representations}
The dialogue context and response can be either visual or textual data. 
We use Vision Transformer ~\citep{dosovitskiy2020image} to learn visual representations of images. Formally, we process the visual image ${v}\in \mathbb{R}^{H \times W \times \mathfrak{C}}$ by dividing it into $N=H W / P^2$ patches ${v}^p \in \mathbb{R}^{N \times\left(P^2 \mathfrak{C}\right)}$, where $\mathfrak{C}$ is the number of channels, $(H, W)$ is the resolution of the input image, and $P$ is the patch resolution. This allows the model to extract meaningful features from the image by considering it as a set of small regions, rather than a single large array of pixels. The image patches are then flattened into vectors and processed by a linear projection using a weight matrix $\mathbf{W}_{V} \in \mathbb{R}^{\left(P^2 \cdot \mathfrak{C}\right) \times E}$ and a position embedding $\mathbf{W}_{V}^{\text{pos}} \in \mathbb{R}^{(N+1) \times E}$, resulting in patch embedding $\bar{v} \in \mathbb{R}^{N \times E}$, where $E$ is the dimension of embedding. 
%The linear projection allows us to obtain a set of dense representations of the image patches that capture relevant information while discarding noise and irrelevant details. 
The position embedding is used to add additional information about the position of the patch in the image. 
Finally, we obtain the visual representations $\boldsymbol{H}_0^v$ after summing patch embeddings and position embeddings.

\paragraph{Textual Representations}
The input text $t \in \mathbb{R}^{\mathfrak{L} \times |O|}$ is embedded into a dense representation $\bar{t} \in \mathbb{R}^{\mathfrak{L} \times E}$ by using a word embedding matrix $\mathbf{W}_T \in \mathbb{R}^{|O| \times E}$ and a position embedding matrix $\mathbf{W}_T^{\text{pos}} \in \mathbb{R}^{(\mathfrak{L}+1) \times E}$, where $|O|$ is the size of the vocabulary, $\mathfrak{L}$ is the length of text, and $E$ is the dimension of embedding. It is noteworthy that we usually concatenate the context with the current utterance to form the final textual input. The textual representations can be denoted as $\boldsymbol{H}_0^t$.

%\subsection{Model Overview}
\subsection{Divide-and-Conquer Pre-training Strategy}
We devise a novel pre-training strategy  in a divide-and-conquer manner. 
Specifically, we first divide the complicated multi-model dialogue into several sub-problems, which can be learned in an easier way. The solutions to the sub-problems are then combined to give a solution to different downstream multi-modal dialogue tasks. 
%Specifically, the solutions to the sub-problems are then combined to give a solution to different downstream multi-modal dialogue tasks. 

\paragraph{Multi-expert Architecture}
PaCE adopts an extension of the standard Transformer, which learns multiple semantic experts instead of a single feedforward network (FFN) as in the original Transformer~\citep{bao2021vlmo}. Concretely, the experts share the information from both textual and visual modalities through a multi-head attention mechanism (MSA), while each expert $\text{FFN}^\text{expert}$ has its own unique parameters to learn a different semantic representation.
Formally, the unique information, which is obtained by switching experts in each block, can be formulated as:
\begin{equation}
\begin{array}{r}
\boldsymbol{H}_l^{\prime}=\text{MSA}\left(\text{LN}\left(\boldsymbol{H}_{l-1}\right)\right)+\boldsymbol{H}_{l-1} \\
\boldsymbol{H}_l^{\text{expert}_k}=\text{FFN}^{\text{expert}_k}\left(\operatorname{LN}\left(\boldsymbol{H}_l^{\prime}\right)\right)+\boldsymbol{H}_l^{\prime}
\end{array}
\end{equation}
where $\boldsymbol{H}_{l-1}$ ($l \in[1, L]$) represents the output representation of the $l$-1 layer and $L$ is the number of Transformer blocks.
$\boldsymbol{H}_l^{\text{expert}_k}$ is the representation of the $k$-th expert. The input representation could be formulated as
$\boldsymbol{H}_0 = [\boldsymbol{H}_0^v, \boldsymbol{H}_0^t]$.
Here, MSA and LN are the standard multi-head self-attention and layer normalization, respectively.\footnote{Due to limited space, we do not elaborate on MSA and LN, and readers can refer to ~\cite{vaswani2017attention} for implementation details.} 

\paragraph{Modality and Capability Experts}
As illustrated in Figure~\ref{method}, we divide the complicated multi-modal dialogue task into five easier sub-problems including \textsc{Caption} modeling, \textsc{Context} modeling, \textsc{Image} modeling, \textsc{Grounding}, and \textsc{Generation}. We design a semantic expert to solve each sub-problem. These five experts can be divided into two categories: modality experts (\textsc{Caption} and \textsc{Image} experts) and capability experts (\textsc{Grounding}, \textsc{Context Modeling} and \textsc{Generation} experts) tailored for multi-modal dialogue. Ultimately, we activate the modality and capability experts in a hierarchical manner, with the bottom $(L-F)$ layers activating only the modality experts and the top $F$ layers activating the capability experts, where $F$ is a pre-defined hyper-parameter.

%Five experts have been developed to meet the varied needs of different tasks. These experts are divided into two categories: modality experts and capability experts. The modality experts include text and image expert, while the capability experts possess proficiency in the three essential capabilities required for multimodal dialog: grounding, context modeling, and generative modeling. Ultimately, we activate the modality and capability experts in a hierarchical manner, with the bottom $(L - F)$ layers activating only the modality experts and the top $F$ layers activating the capability experts, where $F$ is a hyper-parameter.

\paragraph{Experts Combination for Different Tasks}
We propose a progressive cascade pre-training strategy that solves different multi-modal dialogue tasks by adaptively combining the solutions to the sub-problems. 
% \yangmin{Pls add two or three sentences to describe this training strategy.}
We will introduce the details of progressive cascade pre-training in Section 4.3.  

\subsection{Pre-training Objectives}
Our progressive cascade pre-training process consists of three phases, each with a tailored pre-training objective.

\paragraph{Stage I: Image-Text Matching}
In stage I, similar to ViLT~\citep{kim2021vilt}, we use non-dialogue multi-modal data $D_n$ to learn the fundamental inter-modal alignment, and this stage involves only three experts, including the \textsc{Caption} expert,  \textsc{Image} expert and  \textsc{Grounding} expert.
As depicted in Figure~\ref{method}(a), following word and patch embeddings, the text and image are separately processed into text and image representations by specialized \textsc{Caption} and \textsc{Image} experts. These representations are then fused and fed into the \textsc{Grounding} expert, yielding a unified representation of the image and text. We then employ the representation of the `\texttt{[CLS]}' token from the expert output as the input for a binary classification network to predict the alignment between the current text and image. The loss function of image-text matching is defined as:
\begin{equation}
\mathcal{L}_{\mathrm{itm}}=\mathbb{E}_{(V, T) \sim D_n} \mathrm{CE}\left(\boldsymbol{y}_{\mathrm{itm}}, \boldsymbol{p}_{\mathrm{itm}}(V, T)\right)
\end{equation}
In addition to $\mathcal{L}_{\mathrm{itm}}$, we also employ the MLM loss $\mathcal{L}_{\mathrm{mlm}}$ in this stage for  understanding unique textual modality.
Concretely, following the method of BERT, we randomly select tokens in the text sequence and replace them with the \texttt{[MASK]} token. The model is trained to predict these masked tokens using the context of the remaining unmasked tokens and the visual clues. We adopt a masking probability of 15\%. The final output vectors of the masked tokens are then fed into a classifier over the entire text vocabulary, with the training loss being the cross-entropy loss.
\begin{equation}
% \small
\mathcal{L}_{\mathrm{mlm}}=\mathbb{E}_{(V, \hat{T}) \sim {\{D_n \cup D_d\}}} \mathrm{CE}(\boldsymbol{y}_{\mathrm{mask}}, \boldsymbol{p}_{\mathrm{mask}}(V, \hat{T}))
\end{equation}
where $\hat{T}$ is a masked text, $V$ is an original image and $\boldsymbol{p}_{\mathrm{mask}}(V, \hat{T})$ denotes the model's predicted probability for the masked token $\hat{T}$. 
$D_n$ and $D_d$ represent multi-modal non-dialogue and dialogue data, respectively.

The joint loss in stage I can be formulated as:
\begin{equation}
\mathcal{L}_{\mathrm{stage}}^{\mathrm{I}} = \mathcal{L}_{\mathrm{itm}} + \mathcal{L}_{\mathrm{mlm}}
\end{equation}

\paragraph{Stage II: Image-Context Matching}
In stage II, we use multi-modal dialogue data $D_d$ to pre-train PaCE, which aims to model dialogue context for multi-modal dialogue tasks. At this stage,  \textsc{Caption} expert will be activated in addition to the three experts from the first stage. Concretely, in the second stage, the dialogue context $C$ is input to \textsc{Context} expert, the images $V$ are input to \textsc{Image} expert, and the corresponding image captions $T$ are input to  \textsc{Caption} expert. The loss function of  image-context matching is defined as:
\begin{equation}
\mathcal{L}_{\mathrm{icm}}=\mathbb{E}_{(V, T, C) \sim D_d} \mathrm{CE}\left(\boldsymbol{y}_{\mathrm{icm}}, \boldsymbol{p}_{\mathrm{icm}}(V, T, C)\right)
\end{equation}
In addition, we use the \textsc{Caption} expert learned in Stage I as a teacher to facilitate the learning of \textsc{Context} expert.
\begin{equation}
\mathcal{L}_{\mathrm{tca}}= \left\|\boldsymbol{H}_{L-F}^t-\boldsymbol{H}_{L-F}^c\right\|_2^2,
\end{equation}
where $\boldsymbol{H}_{L-F}^t$ and $\boldsymbol{H}_{L-F}^c$ are the output of the $\{L-F\}$th-layer of \textsc{Caption} expert and \textsc{Context} expert, respectively.

Besides, we also employ MLM loss in stage II as defined in stage I, and the joint loss $\mathcal{L}_{\mathrm{stage}}^{\mathrm{II}}$ in stage II could be formulated as:
\begin{equation}
\mathcal{L}_{\mathrm{stage}}^{\mathrm{II}} = \mathcal{L}_{\mathrm{icm}} + \mathcal{L}_{\mathrm{tca}} +  \mathcal{L}_{\mathrm{mlm}}
\end{equation}

\paragraph{Stage III: Generation Modeling}
The third stage aims to enable the model to generate responses. The \textsc{Generation} expert is activated, and the input to this expert is composed of the \textsc{Context} expert and the \textsc{Image} expert. The loss function in stage III is defined as follows:
%\yangmin{why not use text expert?}
\begin{equation}
\mathcal{L}_{\mathrm{stage}}^{\mathrm{III}}=-\sum_{n=1}^N \log \boldsymbol{p}_{\mathrm{rgm}}\left(C_n \mid V, C_{<n}\right)
\end{equation}
Here, we model generative capability by auto-regression, i.e., using past dialogue history $C_{<n}$ and associated images $V$ to predict the current turn $C_n$ of a dialogue.
%\yangmin{Pls define every unknown notation, e.g., $C_n$, $V$, $T$.}

\subsection{Fine-Tuning on Downstream Tasks}
Once the pre-training of PaCE is finished, we perform fine-tuning on specific downstream tasks. Thanks to our divide-and-conquer 
pre-training approach, we can flexibly select different capability experts to solve a specific downstream task. 
Specifically, for understanding tasks, including intent prediction, dialog retrieval, and dialog state tracking, we activate \textsc{Context} expert, \textsc{Image} expert, and \textsc{Grounding} expert. 
For the generation task, i.e. dialog state tracking, and response generation, we activate the \textsc{Context} expert, \textsc{Image} expert, and \textsc{Generation} expert. 

% \yangmin{I think the fine-tuning should describe how to use the pre-training model, e.g., how to combine the experts to accomplish a specific downstream task, instead of just describing the tasks in the current version.}

% \clearpage
\section{Experiments}

\begin{table*}[h]
    \centering
    \resizebox{0.8\textwidth}{!}{
    \begin{tabular}{lllll}
    % \toprule
    \toprule
    \textbf{Task} & \textbf{Dataset} & \textbf{Metric} & \textbf{Previous SOTA} & \textbf{PaCE} \\
    \cmidrule(lr){1-5}
    \cmidrule(lr){1-5}
    % \hline
    \multirow{2}{*}{Multi-Modal Intent Prediction} 
    & PhotoChat & F1-Score & 58.9 (T5-3B)& \textbf{63.8} (\green{\textbf{+4.9}}) \\
    \cmidrule(lr){2-5}
    & MMDialog & F1-score & 75.5 (Divter) & \textbf{77.6} (\green{\textbf{+2.1}}) \\
    \cmidrule(lr){1-5}
    \multirow{2}{*}{Multi-Modal Dialog Retrieval} & PhotoChat (T2I) & R@1 & 10.4 (SCAN) & \textbf{15.2} (\green{\textbf{+4.8}}) \\
    \cmidrule(lr){2-5}
    & Image-Chat (I2T) & R@1 & 50.3 (TransResNet) & \textbf{51.9} (\green{\textbf{+1.6}})\\
    \cmidrule(lr){1-5}
    \multirow{2}{*}{Multi-Modal Dialog State Tracking} & MMConv & Acc. & 18.0 (DS-DST) & \textbf{39.2} (\green{\textbf{+21.2}}) \\
    \cmidrule(lr){2-5}
    & SIMMC2.0 & Act-F1 & 96.3 (BART-large) & \textbf{97.1} (\green{\textbf{+0.8}}) \\
    \cmidrule(lr){1-5}
    \multirow{2}{*}{Multi-Modal Response Generation} & MMConv & Comb. & 32.2 (SimpleTOD) & \textbf{44.7} (\green{\textbf{+12.5}}) \\
    \cmidrule(lr){2-5}
    & SIMMC2.0 & BLEU & 33.1 (BART-large) & \textbf{34.1} (\green{\textbf{+1.0}})  \\
    \bottomrule
    \end{tabular}}
    \caption{
    Experimental results on various multi-modal dialogue benchmarks. We compare PaCE with previous state-of-the-art models, including T5-3B~\citep{raffel2020exploring}, Divter~\citep{feng2022mmdialog}, SCAN~\citep{lee2018stacked}, TransResNet~\citep{shuster2018image}, BART-large~\citep{lewis2019bart} and SimpleTOD~\citep{hosseini2020simple}.
    }
    \label{tab:benchmark}
\end{table*}

\subsection{Downstream Datasets}
To comprehensively evaluate our PaCE, we conduct extensive experiments on seven datasets belonging to four downstream tasks.

\paragraph{Multi-Modal Intent Prediction}
For multi-modal intent prediction,  PhotoChat~\citep{zang2021photochat} and MMDialog~\citep{feng2022mmdialog} are selected as benchmark datasets.
This task aims to identify the specific intent of the user in the multi-modal context. More specifically, it predicts the probability of photo sharing in the upcoming conversation turn.

\paragraph{Multi-Modal Dialog Retrieval} 
For text-to-image retrieval, we select PhotoChat~\citep{zang2021photochat} as our benchmark dataset. It encompasses 12k dialogues, each accompanied by a user photo exchanged during the conversation. The goal of this task is to select the most appropriate photo given the dialog context. For image-to-text retrieval, we select Image-Chat~\citep{shuster2018image} to evaluate our model, which consists of 202k dialogues over 202k images. 
%It requires an AI agent to select the most relevant textual response based on natural, conversational textual data about visual content.

\paragraph{Multi-Modal Dialog State Tracking}
MMConv~\citep{liao2021mmconv} and SIMMC2.0~\citep{kottur2021simmc} datasets provide a good base for carrying out multi-modal dialog state tracking. The MMConv dataset contains 5.1k dialogues collected by enabling multi-modal conversations between human-to-human role-playing pairs under real-life traveling scenarios. In contrast, the SIMMC2.0 corpus includes 11,000 task-oriented dialogs in the shopping domain that are grounded in immersive and photo-realistic contexts.

\paragraph{Multi-Modal Response Generation}
Generating appropriate responses for satisfactory task completion is the ultimate goal of task-oriented dialogue agents. In this task, we selected MMConv~\citep{liao2021mmconv} and SIMMC2.0~\citep{kottur2021simmc} as our benchmark datasets.
\subsection{Experimental Setting}
We use the \textit{bert-base-uncased} tokenizer to tokenize text inputs. We learn the textual embedding-related parameters from scratch, instead of fine-tuning them from pre-trained BERT.
For all experiments, we use AdamW optimizer~\citep{loshchilov2017decoupled} with base learning rate of $10^{-4}$ and weight decay of $10^{-2}$. The learning rate is warmed up for 10\% of the total training steps and is decayed linearly to zero for the rest of the training. We set the total number of the Transformer layers L to 12, with the number of layers F for the top Transformer set to 3.
We initialize the Transformer weights with the pre-trained ViT ~\citep{dosovitskiy2020image}. 
In the pre-training process, we utilize 200K steps, 25K steps, and 10K steps, respectively, for the three stages on 8 NVIDIA A100 GPUs with a batch size of 4,096.
%We apply RandAugment~\citep{cubuk2020randaugment} during the fine-tuning process. For all downstream tasks, we train our model for ten epochs with a batch size of 256.

\subsection{Evaluation Metrics}
 For intent prediction, we adopt the F1 score as the evaluation metric to measure the effectiveness of our model, similar to previous work~\cite{zang2021photochat}. For multi-modal dialog retrieval, we use ranking-based evaluation metrics such as recall \textit{n} at \textit{k} including \textit{R@1}, \textit{R@5} and \textit{R@10} in accordance with prior studies~\citep{zang2021photochat, shuster2018image}. These metrics measure whether the ground-truth textual or visual outputs are ranked among the top $ k \in \{1,\ 5,\ 10\}$ positions among \textit{n} candidate elements. For multi-modal dialogue state tracking, we report \textit{Categorical}, \textit{Non-categorical} and \textit{overall} scores as evaluation metrics following~\cite{liao2021mmconv}. To measure the quality of response generation, we employ BLEU~\cite {papineni2002bleu} as the evaluation metric for SIMMC2.0. For MMConv, we report a combined score (Comb.), which is computed via $(Inform + Success)\times 0.5+BLEU$ as an overall evaluation measure as in ~\cite{mehri2019structured}. 
 %For SIMMC2.0, we report BLEU as same as in ~\citet{kottur2021simmc}.
% \subsection{Benchmark Performance}

% \subsection{Baselines}

\subsection{Quantitative Comparison}
As shown in Figure~\ref{fig:result} and Table~\ref{tab:benchmark}, PaCE  demonstrates state-of-the-art performances across a wide range of multi-modal dialogue tasks. Specifically, we have achieved a significant enhancement on the PhotoChat and MMConv dataset, with an improvement of 4.8 points in multi-modal dialog retrieval and 21.2 points in multi-modal dialog state tracking, respectively. It is worth noting that PaCE has a total parameter count of 338 million. In addition, since some experts may be idle during the execution of specific downstream tasks, the parameter size will further decrease for specific downstream tasks. Below, we provide a detailed analysis of the results for each sub-task dataset.

\paragraph{Multi-Modal Intent Prediction}
For the PhotoChat dataset, we report the performances of strong baselines as in ~\citep{zang2021photochat}, including ALBERT-base~\citep{lan2019albert}, BERT~\citep{devlin2018bert}, T5-base, and T5-3B~\citep{raffel2020exploring}. For the MMDialog dataset, we adopt DE++,  Divter~\citep{feng2022mmdialog}, and ViLT~\citep{kim2021vilt} as our baseline models. %where are both designed based on CLIP~\citep{radford2021learning}. In addition, we also conduct experiments on a traditional VLM model, i.e. ViLT~\citep{kim2021vilt}. 
As shown in Table~\ref{tab:ap-intent-photochat},
although some models such as T5-3B are much larger than ours, our model still achieves the best performance on all evaluation metrics. 
% This verifies the effectiveness of our model. 
\begin{table}[t!]
    \centering
    \resizebox{0.48\textwidth}{!}{
    \begin{tabular}{llcc|lc}
    \toprule
    \multicolumn{4}{c|}{\textbf{PhotoChat}} & \multicolumn{2}{c}{\textbf{MMDialog}} \\
    \cmidrule(lr){1-4}
    \cmidrule(lr){5-6}
    \textbf{Model} & \textbf{F1} & \textbf{Precision} & \textbf{Recall} & \textbf{Model} & \textbf{F1}\\
    % \cmidrule(lr){1-5}
    % \hline\hline
    \midrule
    \midrule
    ALBERT-base & 52.2 & 44.8& 62.7 & DE++ & 59.0 \\
    BERT-base & 53.2 & 56.1 & 50.6  & Divter & \underline{75.5} \\
    T5-base & 58.1 & \underline{58.2} & 57.9 &-&-\\
    T5-3B & \underline{58.9} & 54.1 & \underline{64.6}&-&- \\
    ViLT & 52.4 & 55.4& 58.9 & ViLT & 55.8 \\
    % \cmidrule(lr){1-5}	
    \rowcolor{gray!20} \textbf{PaCE} & \textbf{63.8} & \textbf{63.3} & \textbf{68.0} & \textbf{PaCE} & \textbf{77.6}\\ 
    
    \bottomrule
    \end{tabular}}
    \caption{
    Multi-modal intent prediction results on PhotoChat and MMDialog.
    }
    \label{tab:ap-intent-photochat}
\end{table}

% \paragraph{intent}
% \paragraph{...}

\paragraph{Multi-Modal Dialog Retrieval}
For PhotoChat, we compare PaCE with strong baselines reported in~\citep{zang2021photochat}, including BM25~\citep{robertson2009probabilistic}, 
DE$^*$~\citep{zang2021photochat},
VSE++~\citep{faghri2017vse++} and SCAN~\citep{lee2018stacked}. We also adapted VLMo~\citep{bao2021vlmo} and ViLT~\citep{kim2021vilt} to perform multi-modal dialog retrieval.
The results on PhotoChat are reported in Table~\ref{tab:ap-retrieval-photochat}, PaCE achieves substantially better performance than the best performing baselines. 
For Image-Chat, we compare PaCE with TransResNet152~\citep{liao2021mmconv}, VLMo and ViLT, and report baseline results as in Table~\ref{tab:ap-image-chat}. 
%Experimental results show that 
PaCE achieves the best results for image-to-text dialog retrieval with 3.0 improvement in terms of Sum.

\begin{table}[t!]
    \centering
    \small
    \resizebox{0.45\textwidth}{!}{
    \begin{tabular}{ccccc}
    \toprule
    \textbf{Model} & \textbf{R@1} & \textbf{R@5} & \textbf{R@10} & \textbf{Sum(R@1,5,10)} \\
    % \cmidrule(lr){1-5}
    % \cmidrule(lr){1-5}
    % \hline\hline
    \midrule
    BM25 & 6.6&15.4&23.0&45.0\\
    DE$^*$ &9.0 & 26.4 & 35.7 & 71.1 \\
    VSE++ & 10.2 & 25.4 & 34.2 & 69.8 \\
    SCAN & 10.4 & 27.0& 37.1 & 74.5\\
    VLMo & \underline{13.8}& \underline{30.0} & \underline{39.4} & \underline{83.2}\\
    ViLT & 11.5	& 25.6&33.8	& 71.0 \\
    % \cmidrule(lr){1-5}
    \rowcolor{gray!20} \textbf{PaCE} & \textbf{15.2} & \textbf{36.7} & \textbf{49.6} & \textbf{101.5} \\
    \bottomrule
    \end{tabular}}
    \caption{
    Multi-modal dialogue retrieval on PhotoChat.
    }
    \label{tab:ap-retrieval-photochat}
\end{table}

\begin{table}[t!]
    \centering
    \small
    \resizebox{0.45\textwidth}{!}{
    \begin{tabular}{lccc}
    \toprule
    \textbf{Model} & \textbf{R@1} & \textbf{R@5} & \textbf{Sum(R@1,5)} \\
    % \cmidrule(lr){1-5}
    % \cmidrule(lr){1-5}
    % \hline\hline
    \midrule
    TransResNet152 & 40.6 & 67.2& 107.8\\
    TransResNet152-IG-3.5B & \underline{50.3} & \underline{75.4} & \underline{125.7}  \\
    VLMo & 46.8 & 67.5 & 114.3 \\
    ViLT & 48.4 & 70.0 & 118.4 \\
    % \cmidrule(lr){1-5}
    \rowcolor{gray!20} \textbf{PaCE} & \textbf{51.9} & \textbf{76.8} & \textbf{128.7}\\
    
    \bottomrule
    \end{tabular}}
    \caption{
    Multi-modal dialog retrieval on Image-Chat.
    }
    \label{tab:ap-image-chat}
\end{table}

\vspace{-0.3mm}
\paragraph{Multi-Modal Dialog State Tracking}
For MMConv dataset, we compare PaCE with DS-DST\citep{zhang2019find}; for SIMMC2.0 dataset, we compare PaCE with GPT-2~\citep{radford2019language}, MTN~\citep{le2019multimodal}, BART-large and BART-base~\citep{lewis2019bart}. The results on MMConv and SIMMC2.0 are reported in Table~\ref{tab:ap-mmconvdst} and Table~\ref{tab:ap-simmcdst}, respectively.  
PaCE can achieve the best results on most of the evaluation metrics. 
Notably, we observed that the PaCE achieves competitive results at smaller parameter scales than previous SOTA in SIMMC2.0 slot F1.

\begin{table}[t!]
    \centering
    \small
    \resizebox{0.45\textwidth}{!}{
    \begin{tabular}{lccc}
    \toprule
    \textbf{Model} & \textbf{Categorical} & \textbf{Non-categorical} & \textbf{Overall} \\
    % \cmidrule(lr){1-5}
    % \cmidrule(lr){1-5}
    % \hline\hline
    \midrule
    DS-DST & 91.0 & 23.0 & 18.0\\
    % \cmidrule(lr){1-5}0.922	0.434	0.390
    \rowcolor{gray!20} \textbf{PaCE} & \textbf{92.2} & \textbf{43.4} & \textbf{39.2}\\
    \bottomrule
    \end{tabular}}
    \caption{
    Multi-modal dialog state tracking performances on MMConv.
    }
    \label{tab:ap-mmconvdst}
\end{table}

\begin{table}[t!]
    \centering
    \small
    \resizebox{0.45\textwidth}{!}{
    \begin{tabular}{lcc|c}
    \toprule
    \multicolumn{3}{c|}{\textbf{Dialog State Tracking}} & \multicolumn{1}{c}{\textbf{Dialog Generation}} \\
    \midrule
    \textbf{Model} & \textbf{Slot F1} & \textbf{Act. F1} & \textbf{BLEU}\\
    % \cmidrule(lr){1-5}
    % \cmidrule(lr){1-5}
    % \hline\hline
    \midrule
    GPT-2 & 81.7 & 94.5 & 19.2 \\
    MTN & 76.7 & 93.4 & 21.7 \\
    BART-large & \textbf{88.3} & \underline{96.3} &\underline{33.1}\\
    BART-base & 82.0 & 95.2 & 29.4 \\
    % \cmidrule(lr){1-5}0.922	0.434	0.390
    \rowcolor{gray!20} \textbf{PaCE} & \underline{87.0} & \textbf{97.1}& \textbf{34.1}\\
    \bottomrule
    \end{tabular}}
    \caption{
     Multi-modal dialog state tracking on SIMMC2.0. The evaluation metrics Slot F1 and Act. F1 are used to evaluate the dialog state tracking task, while BLEU is adopted for evaluating response generation. 
    }
    \label{tab:ap-simmcdst}
\end{table}

\paragraph{Multi-Modal Response Generation}
For the response generation task, we conduct experiments on SIMMC2.0 and MMConv datasets.
For MMConv, we adopt the strong baseline SimpleTOD~\citep{hosseini2020simple} implemented by ~\citep{liao2021mmconv}. We summarize the experimental results of  SIMMC2.0 and MMConv in Table~\ref{tab:ap-simmcdst} and Table~\ref{tab:ap-mmconvrg}, verifying the effectiveness of our model in both discriminative and generative tasks.  
%\yangmin{I think the results in Table 7 should be re-arranged. Differentiate thee dialog state tracking and generation.}

\begin{table}[t!]
    \centering
    \small
    \resizebox{0.45\textwidth}{!}{
    \begin{tabular}{lcccc}
    \toprule
    \textbf{Model} & \textbf{Inform} & \textbf{Success} & \textbf{BLEU} & \textbf{Comb.} \\
    % \cmidrule(lr){1-5}
    % \cmidrule(lr){1-5}
    % \hline\hline
    \midrule
    SimpleTOD & 14.6 & 9.2& 20.3 & 32.2\\
    % \cmidrule(lr){1-5}0.922	0.434	0.390
    \rowcolor{gray!20} \textbf{PaCE} & \textbf{34.5} & \textbf{13.9} & \textbf{22.0} & \textbf{44.7}\\
    \bottomrule
    \end{tabular}}
    \caption{
    Multi-modal response generation performances on MMConv.
    }
    \label{tab:ap-mmconvrg}
\end{table}

\subsection{Ablation Study}
\paragraph{Effectiveness of Pre-training Objectives}
To evaluate the effectiveness of each stage of pre-training, we conduct an ablation study by removing Stage I pre-training~(PaCE$_{\rm w/o~\footnotesize{\mathcal{L}_{\rm stage}^{\rm I}}}$), removing Stage II pre-training (PaCE$_{\rm w/o~\footnotesize{\mathcal{L}_{\rm stage}^{\rm II}}}$), removing Stage III pre-training (PaCE$_{\rm w/o~\footnotesize{\mathcal{L}_{\rm stage}^{\rm III}}}$), and removing both Stage II and Stage III (PaCE$_{\rm only~\footnotesize{\mathcal{L}_{\rm stage}^{\rm I}}}$). For a fair comparison, the experimental setup of the ablation study is consistent with that of the primary experiments, utilizing the same hyper-parameters and downstream fine-tuning strategy. The ablation test results on PhotoChat and Image-Chat are provided in Table \ref{tab:ab-stage1}. We can observe that  image-text matching (Stage I) and image-context matching (Stage II) play the most important role in PaCE.
This is within our expectation since Stage I and Stage II are the basis of the latter generation modeling (Stage III). It is no surprise that combining all three stages achieves the best performance on the experimental datasets.
We also investigate the impact of $\mathcal{L}_{\rm tca}$ by removing it from Stage II pre-training (denoted as PaCE$_{\rm w/o~\mathcal{L}_{\rm tca}}$). We can observe that
$\mathcal{L}_{\rm tca}$ has a significant impact on the performance of PaCE in Stage II pre-training.

\paragraph{Effectiveness of Pre-training Data}
In addition, we also conduct an ablation study to verify the impact of different pre-training data on PhotoChat and Image-Chat datasets. We define the models that only use MultiNonDialog and MultiDialog for pre-training as PaCE$_{\rm only~\rm MultiNonDialog}$ and PaCE$_{\rm only~\rm MultiDialog}$, respectively. The ablation test results on PhotoChat and Image-Chat are provided in Table~\ref{tab:ab-data}. We can observe that both MultiNonDialog and MultiDialog pre-training corpora contribute great performance improvement to PaCE. This is within our expectation since the MultiNonDialog data helps our model learn impressive image-text representations and their alignment, while the MultiDialog data encourages PaCE to capture the dialog context information.
\begin{table}[h]
    \centering
    \resizebox{0.48\textwidth}{!}{
    \begin{tabular}{llclc}
    \toprule
    \multirow{2}{*}{\textbf{Model}} & \multicolumn{2}{c}{\textbf{PhotoChat}} & \multicolumn{2}{c}{\textbf{Image-Chat}} \\
    \cmidrule(lr){2-3}
    \cmidrule(lr){4-5}
    
     & \textbf{R@1} & \textbf{Sum(R@1,5,10)} & \textbf{R@1} & \textbf{Sum(R@1,5)}\\
    \midrule
    % \cmidrule(lr){1-5}
    % \hline\hline
    \rowcolor{gray!20} PaCE & \textbf{15.2} & \textbf{101.5} & \textbf{51.9} & \textbf{128.7} \\
    
    PaCE$_{\rm w/o~\mathcal{L}_{\rm stage}^{\rm I}}$ & 10.7 & 74.3 & 46.5 & 117.8 \\
    PaCE$_{\rm w/o~\mathcal{L}_{\rm stage}^{\rm II}}$ & 12.0 & 74.8 & 48.5 & 119.5 \\
    PaCE$_{\rm w/o~\mathcal{L}_{\rm stage}^{\rm III}}$ & 15.0 & 100.8 & 51.2 & 127.3 \\
    PaCE$_{w/o~\mathcal{L}_{\mathrm{tca}}}$ & 13.2 & 95.9 & 49.7 & 125.6\\
    \bottomrule
    \end{tabular}
    }
    \caption{
     Ablation test results on the multi-modal dialog retrieval task by using different pre-training objectives. 
    }
    \label{tab:ab-stage1}
\end{table}
\begin{table}[h]
    \centering
    \resizebox{0.48\textwidth}{!}{
    \begin{tabular}{llclc}
    \toprule
    \multirow{2}{*}{\textbf{Model}} & \multicolumn{2}{c}{\textbf{PhotoChat}} & \multicolumn{2}{c}{\textbf{Image-Chat}} \\
    \cmidrule(lr){2-3}
    \cmidrule(lr){4-5}
    
     & \textbf{R@1} & \textbf{Sum(R@1,5,10)} & \textbf{R@1} & \textbf{Sum(R@1,5)}\\
    \midrule
    % \cmidrule(lr){1-5}
    % \hline\hline
    \rowcolor{gray!20} PaCE & \textbf{15.2} & \textbf{101.5} & \textbf{51.9} & \textbf{128.7} \\
    PaCE$_{\rm only~\rm MultiNonDialog}$ & 10.9 & 73.6 & 47.1 & 116.9 \\	 
    PaCE$_{\rm only~\rm MultiDialog}$ & 10.7 & 74.3 & 46.2 & 117.3 \\
    \bottomrule
    \end{tabular}
    }
    \caption{
    Ablation test results on the multi-modal dialog retrieval task by using different  pre-training data.
    }
    \label{tab:ab-data}
\end{table}

\subsection{Case Study}
\begin{figure}[t]
    \small
    \centering
    \includegraphics[width=\columnwidth]{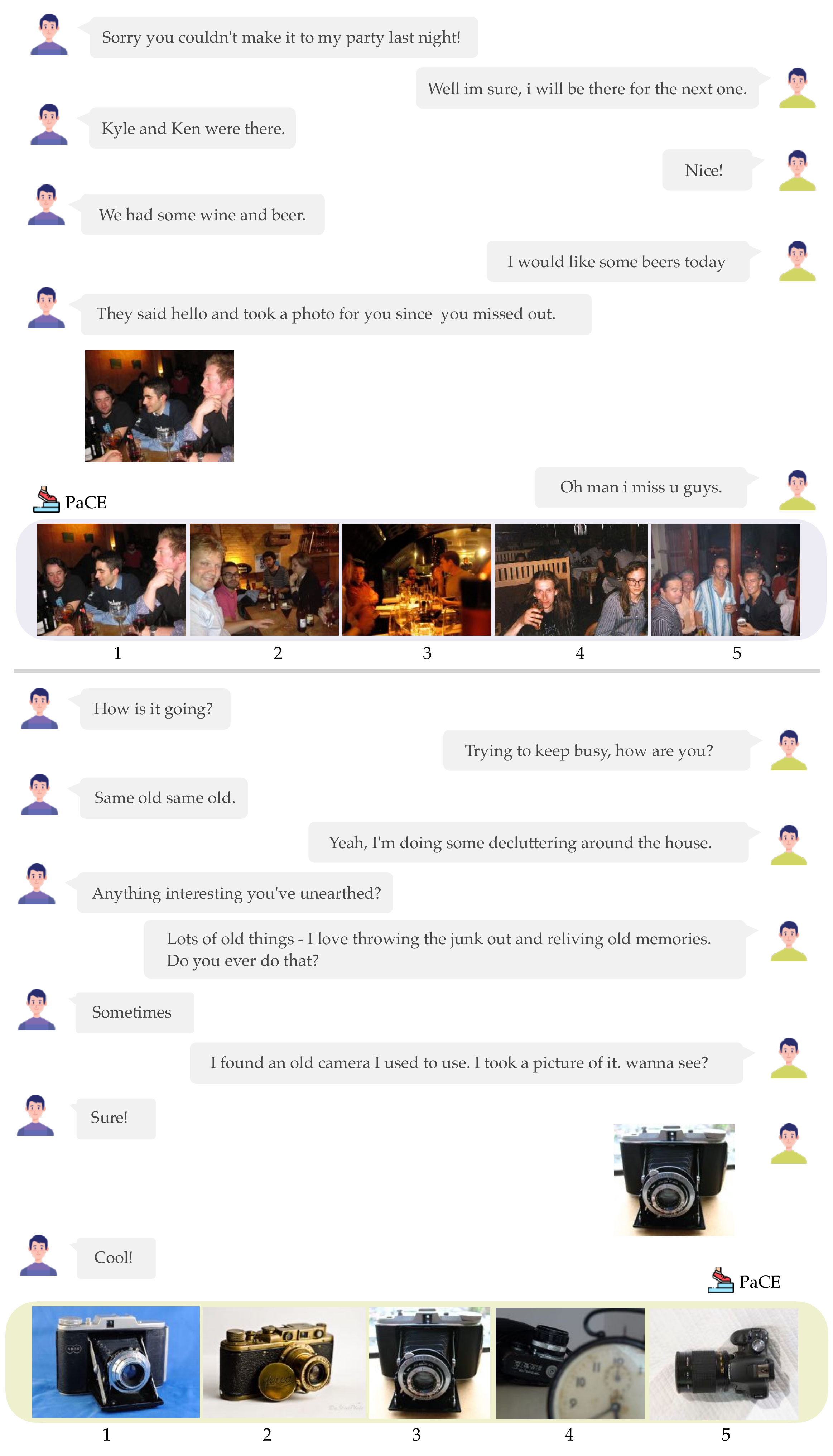}
    \caption{Two cases on the PhotoChat test set. For each dialog query, we show the top-5 ranked images from left to right.}
    \label{casephotochat1}
\end{figure}

\begin{figure}[h]
    \small
    \centering
    \includegraphics[width=0.9\columnwidth]{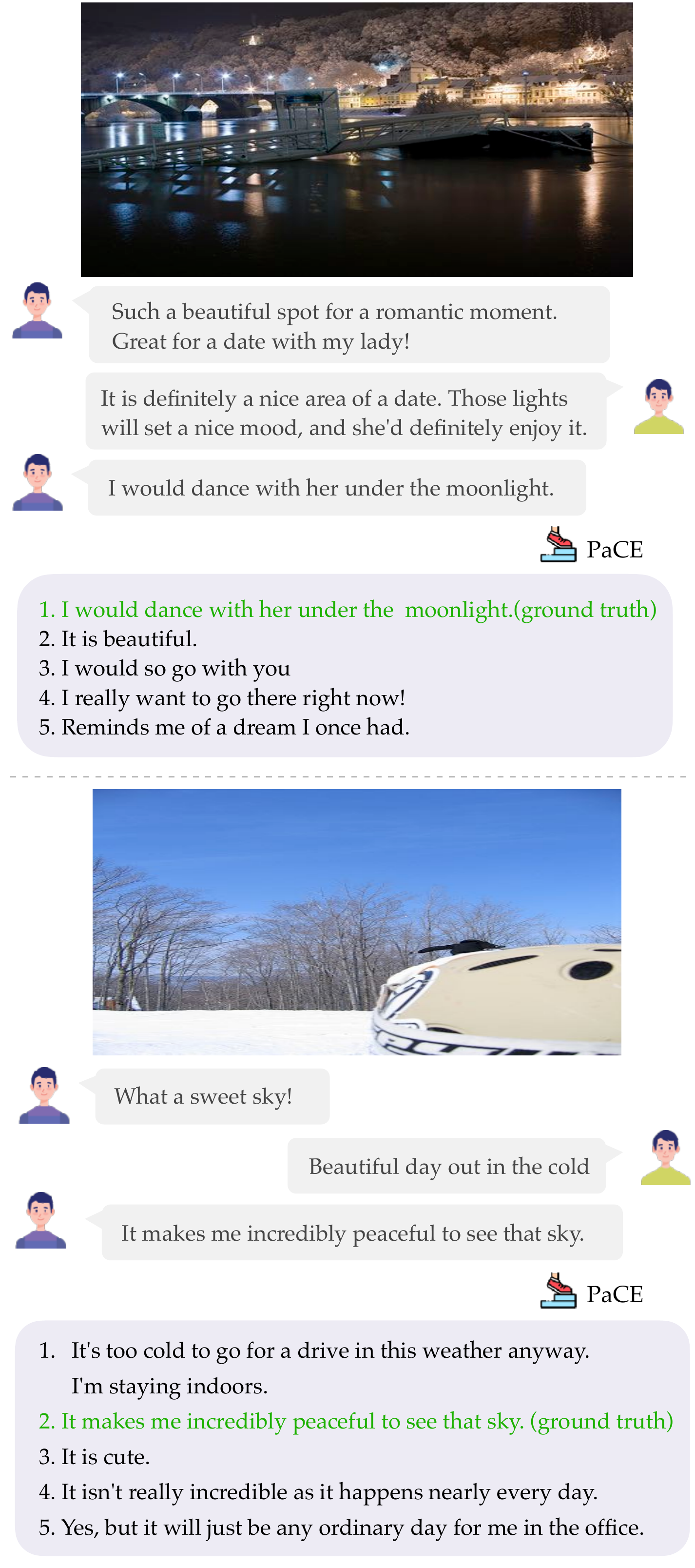}
    \caption{Two cases on the Image-Chat test set. For each dialogue query, we show the top-5 ranked response from top to down.}
    \label{caseimage}
\end{figure}

To evaluate PaCE qualitatively, we choose two exemplary conversations from PhotoChat and Image-Chat test sets, and illustrate the retrieved responses by PaCE in Figure~\ref{casephotochat1} and Figure~\ref{caseimage}. Our PaCE model can retrieve highly relevant candidates to the conversation scenario. For the text-to-image (T2I) retrieval task, since the candidate images could be quite similar, it is challenging to retrieve the exact ground-truth image from the candidates. Although PaCE may not obtain the ground-truth image, we can still obtain the relevant candidate images.  

% \clearpage

\vspace{4mm}
\section{Conclusion}
In this paper, we proposed PaCE, a unified, structured, compositional multi-modal dialogue pre-training framework, which adopted a divide-and-conquer strategy. We first break down the complicated multi-modal dialogue generation task into several sub-capabilities, which could be learned in an easier way. Then, the solutions to the sub-capabilities were combined to obtain an effective and efficient solution to each downstream multi-modal dialogue task. Experimental results on eight benchmark datasets demonstrated that PaCE achieved new state-of-the-art performances.

% \subsection{Citations}
\section*{Discussion}
PaCE adopts a flexible model structure that decomposes complex multimodal dialogues into basic sub-capabilities. As a result, it can be trained progressively on different data and exhibits excellent expandability, making it applicable to new tasks. An additional advantage is that it aligns well with various attempts to enhance performance in terms of interpretability. However, we believe that there are still many aspects of PACE that are worth exploring. First is the exploration of incorporating 
additional modalities and investigating whether the self-attention layer can effectively handle a broader range of modalities for a unified representation. Another aspect worth exploring is the development of a more efficient approach for adapting multi-modal models to diverse downstream applications, eliminating the necessity to fine-tune all parameters of the model. Furthermore, given the substantial variations in the model networks employed for text generation and image generation in contemporary research, exploring the integration of multi-modal generation into a unified framework is a worthwhile endeavor.
% \clearpage
\section*{Limitations}
To better analyze the limitations of PaCE, we carry out an analysis of the errors made by PaCE on the PhotoChat and SIMMC2.0 test sets. We reveal several reasons for the errors, which can be divided into the following categories. \textbf{First}, since there are many similar images in the datasets, PaCE  fail to distinguish some gold image from similar candidates. This may be because we do not design an explicit fine-grained reasoning module to capture the details of images and texts. For example, for the context mentions ``\textit{I and my dad both have a camera}'', our model can capture the entity ``\textit{camera}'', but fails to reason the fact that there should be two cameras. One possible solution is to introduce a  deep reasoning and comprehension strategy to empower the model with excellent reasoning ability.
\textbf{Second}, due to the lack of fine-grained structural understanding of the images, the sentences generated by PaCE suffer from identifying the relative positions of entities. For example, PaCE may have difficulties recognizing the fact that the right side of a yellow shirt is black pants. This issue is particularly severe in SIMMC as there are many entities in the pictures and spatial descriptions of entities in the responses. 
One possible idea is to extract the relative positions of objects mentioned in the conversation as auxiliary data to guide the model's generation.

\section*{Acknowledgements}
Min Yang was partially supported by the National Key Research and Development Program of China (2022YFF0902100), Shenzhen Science and Technology Innovation Program (KQTD20190929172835662), Shenzhen Basic Research Foundation (JCYJ20210324115614039 and JCYJ20200109113441941), and NSFC (no. 92270122). This work was supported by Alibaba Group through Alibaba Innovative Research Program.

%\section*{Acknowledgements}
% \bibliography{anthology,custom}
\bibliography{custom}
\bibliographystyle{acl_natbib}

% \clearpage

\appendix

%\section{Appendix}
%\label{sec:appendix}

\appendix
\newpage

\end{document}